\documentclass[conference,a4paper]{IEEEtran}
\usepackage{cite}
\usepackage[pdftex]{graphicx}
\graphicspath{{./img/}}
\usepackage[tight,footnotesize]{subfigure}
\usepackage[cmex10]{amsmath}
\usepackage{amssymb}
\usepackage{amsfonts}
\usepackage{bm}
\usepackage{algorithmic}

\usepackage{array}
\usepackage{bigstrut}
\usepackage{booktabs}
\usepackage{multirow}
\usepackage{multicol}

\usepackage{breqn}

\def\eg{\emph{e.g }}
\def\ie{\emph{i.e }}

\newcommand{\argmax}{\operatornamewithlimits{arg\ max}}

\usepackage{color, colortbl}

\begin{document}
%
\title{A Graph Transduction Game for Multi-target Tracking}

\author{\IEEEauthorblockN{Tewodros Mulugeta Dagnew\IEEEauthorrefmark{1}, 
Dalia Coppi\IEEEauthorrefmark{2}, Marcello Pelillo\IEEEauthorrefmark{1}, Rita Cucchiara\IEEEauthorrefmark{2}}
\IEEEauthorblockA{\IEEEauthorrefmark{1}DAIS - Ca\'\ Foscari University\\
Venezia, Italy\\
Email: dagnew.tewodros@gmail.com, pelillo@dsi.unive.it}
\IEEEauthorblockA{\IEEEauthorrefmark{2}DIEF - University of Modena and Reggio Emilia\\
Modena, Italy\\
Email: dalia.coppi@unimore.it, rita.cucchiara@unimore.it}}
\maketitle

\begin{abstract}
Semi-supervised learning is a popular class of techniques to learn from labeled and unlabeled data. 
The paper proposes an application of a recently proposed approach of graph transduction that exploits game theoretic notions to the problem of multiple people tracking. 
Within the proposed framework, targets are considered as players of a multi-player non-cooperative game. The equilibria of the game is considered as a consistent labeling solution and thus an estimation of the target association in the sequence of frames.
Patches of persons are extracted from the video frames using a HOG based detector and their similarity is modeled using distances among their covariance matrices. 
The solution we propose achieves satisfactory results on video surveillance datasets. The experiments show the robustness of the method even with a heavy unbalance between the number of labeled and unlabeled input patches.    
\end{abstract}


%
\IEEEpeerreviewmaketitle

\section{Introduction and related work}
\label{sec:intro}

Semisupervised learning (SSL) has recently gained a considerable interest in the machine learning and computer vision communities. 
The reason of the diffusion of SSL method can be easily explained observing that in many real world applications unlabelled data are easy to find, while labelled data are expensive and difficult to retrieve. Typical examples can be found in digital forensics and video surveillance large amount of images and video are readily available, but, without effective automatic tools, corresponding annotation and analysis usually requires the human intervention. Data coming from surveillance cameras need to be examined in order to identify targets that may be captured by different cameras or in different time intervals, therefore in these contexts, many problems can be interpreted as SSL problems, where the initial labelled samples are available (\eg patches of the selected target, some initial frames of the video) in vast footage of unlabelled data.



A comprehensive survey on SSL can be found in \cite{zhu06} and, among the others, graph based algorithms have a relevant role. In such methods labeled and unlabelled input data are modeled as nodes of undirected graphs whose edges represent the similarity among data points. Labels for unlabeled instances are estimated by propagating the information available at labeled nodes to unlabeled nodes. 
The underline idea is that point close to each other (\ie point with high similarity) have the same label. Labels are usually estimated in regularization frameworks where the aim is to minimize a function on the undirected graph \cite{joachims03,zhou05 }. Graph based approaches have been effectively used to solve several problems in computer vision and pattern recognition \cite{tesfayeapplications}, \cite{zemene2016simultaneous}, \cite{ZemeneAP16}, \cite{ZemeneAP17}, \cite{eyasuPAMIgeoloc}, \cite{zemene2015dominant},\cite{YonieyaPAMICDSC_tracker} and also in medical imaging domain \cite{dagnew2017learning}.
The main advantage of graph methods is to be non-parametric and transductive in nature (the solution is a cut of the graph, thus a label for unlabelled input point and not a general classification function as in the case of inductive methods \cite{zhu06}). 

Referring to the video forensics and surveillance fields and considering the tracking problem in its common definition of "problem to estimate the location of target objects in a sequence of images starting from an initial detection", many different approaches have been proposed. In \cite{smeulders13} a large experimental survey of various tracking approaches is presented evaluating the suitability of each approach in different situations and with different constraints (\eg assumptions on the background, on the motion model, on the occlusions, etc.). 
The survey also suggests how the problem of people tracking can be viewed as a data association problem, and in particular as a semi-supervised classification, where some examples of the people to track are initially detected.  

The transductive learning paradigm has already been exploited to solve tracking and re-identification problems. The seminal work of \cite{wu10} introduced the TL as a solution to the severe variation of the models in color tracking. They fitted the TL problem into an EM frameworks to estimate the pixel labels in hand and face color tracking. \cite{zha10} proposed an on-line single target tracking using graph transduction applied to faces and cars. \cite{coppi11} proposed an on-line single target tracking and re-identification method based on a graph based formulation of the TL problem. They enforced the knowledge encoded in the labelled instances introducing an update strategy to avoid drift in the tracking and to allow re-identification in case of occlusions. In \cite{YonET2016} , \cite{AmiShaECCV12},\cite{YonieyaPAMICDSC_tracker}, \cite{gmmcp, tesfaye2014multi} they consider all the pairwise relationships between detection responses in a temporal sliding window, which is used as an input to their optimization based on fully-connected edge-weighted graph.

In this paper we propose to exploit the graph transduction to track multiple people in videos. 
To our knowledge this is the first application of transductive learning to multiple target tracking.
Given few labelled patches of the people to follow we propose a formulation of the multitarget tracking as a graph transduction problem. We exploit the formulation of \cite{erdem12} based on a game theoretic framework and we prove its reliability in solving a real world problem.
We propose an applications able to work off-line on pre-recorded video streams.
Similarly to \cite{coppi11} we describe people patches with covariance matrices and we build the similarity graph using distance among covariance in Riemannian manifolds. 

The rest of the paper is organized as follows: Section \ref{sec:graphTransduction} summarizes the algorithmic framework proposed in \cite{erdem12} and subsequently Section \ref{sec:tracking} details how we propose to apply it in the context of multiple people tracking. In Section \ref{sec:experiments} the performance of the application are evaluated on video surveillance datasets, finally Section \ref{sec:conclusion} gives a glimpse on future works and possible extensions.

\section{Graph Transduction Game}
\label{sec:graphTransduction}
The theoretical formulation of the \emph{Graph Transduction Game} (GTG) has been recently introduced in \cite{erdem12}. Starting from the basis of the transductive learning on undirected graph, they build a solution in which the label estimation is based on game-theoretic notions, in contrast to common solution based on the spectrum of the graph.
Precisely the graph transduction is formulated as a non-cooperative multiplayer game and the labelling correspond to the Nash equilibria.

For notions about multi-player games we refer to \cite{weinbull95}, for completeness we only underline that the main idea is that a \emph{game} is modelled as a strategic interaction among \emph{players} where the goal of each player is to maximize its own \emph{payoff} by choosing the best action (\emph{pure strategy}) to play.
The \emph{graph transduction game} in \cite{erdem12} is formulated assuming that players $i\in\mathcal{I}$ participating in the game corresponds to a particular point in a data set $\mathcal{D}=\{{d}_1,\dots,{d}_n\}$ and that the set of strategy among whom the players can choose is $S_i=\{1,\dots,c\}$. Each strategy $S_i$ expresses a certain hypothesis about its membership to a class and $c$ is the total number of classes (\ie the the mixed strategy profile of each player $i\in\mathcal{I}$ lies in the \mbox{$c$-dimensional} simplex $\Delta_i$). 

Since the problem is a problem of SSL, the players belongs to two disjoint groups: those which already have knowledge of their membership, referred to as \emph{labelled players} and denoted with the symbol $\mathcal{I}_\ell$, and those which do not have any idea about their membership at the beginning of the game, which are hence called \emph{unlabelled players} and correspondingly denoted with$\mathcal{I}_u$. 
The labelled players $\mathcal{I}_\ell=\{\mathcal{I}_{\ell\mid 1},\dots,\mathcal{I}_{\ell\mid c}\}$ do not need to maximise their payoff since they always play their already chosen $k^{th}$ pure strategy where $k = {1, \ldots, c}$. The transduction game can be easily reduced to a game with only unlabelled players $\mathcal{I}_u$ that need to find their mixed strategy $e_i^k\in\Delta_i$ and the fixed strategies of labelled players $\mathcal{I}_\ell$ act as bias over the choices of unlabelled players. 
For the Nash equilibrium theorem \cite{nash51} the GTG always has equilibrium in mixed strategies that corresponds to a steady state where each player plays a strategy that could yield the highest payoff when the strategies of the remaining players are kept fixed, and it provides us a globally consistent labeling of the data set. Once an equilibrium is reached, the label of a data point (player) $i$ is simply given by the strategy with the highest probability in the equilibrium mixed strategy of player $i$  as 
\begin{equation}
\phi_i=\argmax{h = 1 \ldots c}x_{ih},
\end{equation}
thereby yielding a crisp classification.

Similarly to other graph transduction methods the data are represented with an undirected graph $\mathcal{G}=(\mathcal{V},\mathcal{E})$ where $\mathcal{V}$ is the set of nodes representing both labelled and unlabelled points, and $\mathcal{E}$ are the edges weighted with an adjacency matrix $W=(w_{ij})$. Being the solution considered as the equilibrium in a non-cooperative game, the adjacency matrix $W$ is used to compute the pay-off between players. And being the game considered as an instance belonging to the class of \emph{polymatrix games} \cite{quintas89,howson72} where players are nodes of a graph and every edge denote a two-player game between corresponding pair of players, the partial pay-off matrix between two players $i$ and $j$ is computed as $A_{ij}=w_{ij}\times I_{c}$ where $I_{c}$ is the identity matrix of size $c$. 
The pay-offs are then computed as $u_i(e_i^h)=\sum_{j=1}^n (A_{ij} x_j)_h$ and $u_i(x)=\sum_{j=1}^n x_i^T A_{ij} x_j$.

The Nash equilibria, thus the labelling for unlabelled points, is computed using the \emph{evolutionary approach} \cite{daskalakis06,daskalakis11}. 
The dynamic interpretation of Nash equilibria through the evolutionary approach imagines that the game is played repeatedly, generation after generation, during which a selection process acts on the multi-population of strategies, thereby resulting in the evolution of the fittest strategies. The particular class of dynamics used in this are the so called \emph{imitation dynamics} given by
\begin{dmath}
\dot{x}_{ih}=x_{ih}\left[\sum_{l\in S_i}x_{il}\Big(\phi_i\left[u_i\left(e_i^h-e_i^l,x_{-i}\right)\right]- \phi_i\left[u_i\left(e_i^l-e_i^h,x_{-i}\right)\right]\Big)\right] 
\end{dmath}
where the dot signifies derivative w.r.t. time and $\phi_i(u_i)$ is a strictly increasing function of $u_i$. The multi-population version of the replicator dynamics is obtained when $\phi_i$ is taken as the identity function, \textit{i.e.} \mbox{$\phi_i(u_i)=u_i$}, as:
\begin{equation}
\dot{x}_{ih} = x_{ih}\left(u_i(e_i^h,x_{-i})-u_i(x)\right)
\label{eq:repdyncont}
\end{equation}
\cite{erdem12} demonstrate how in both the discrete and continuous time version of the imitation dynamics the fixed points of Eq. \ref{eq:repdyncont} are Nash equilibria.

\section{Multitarget Tracking}
\label{sec:tracking}
\begin{figure*}[tb]
   \includegraphics[width=\textwidth]{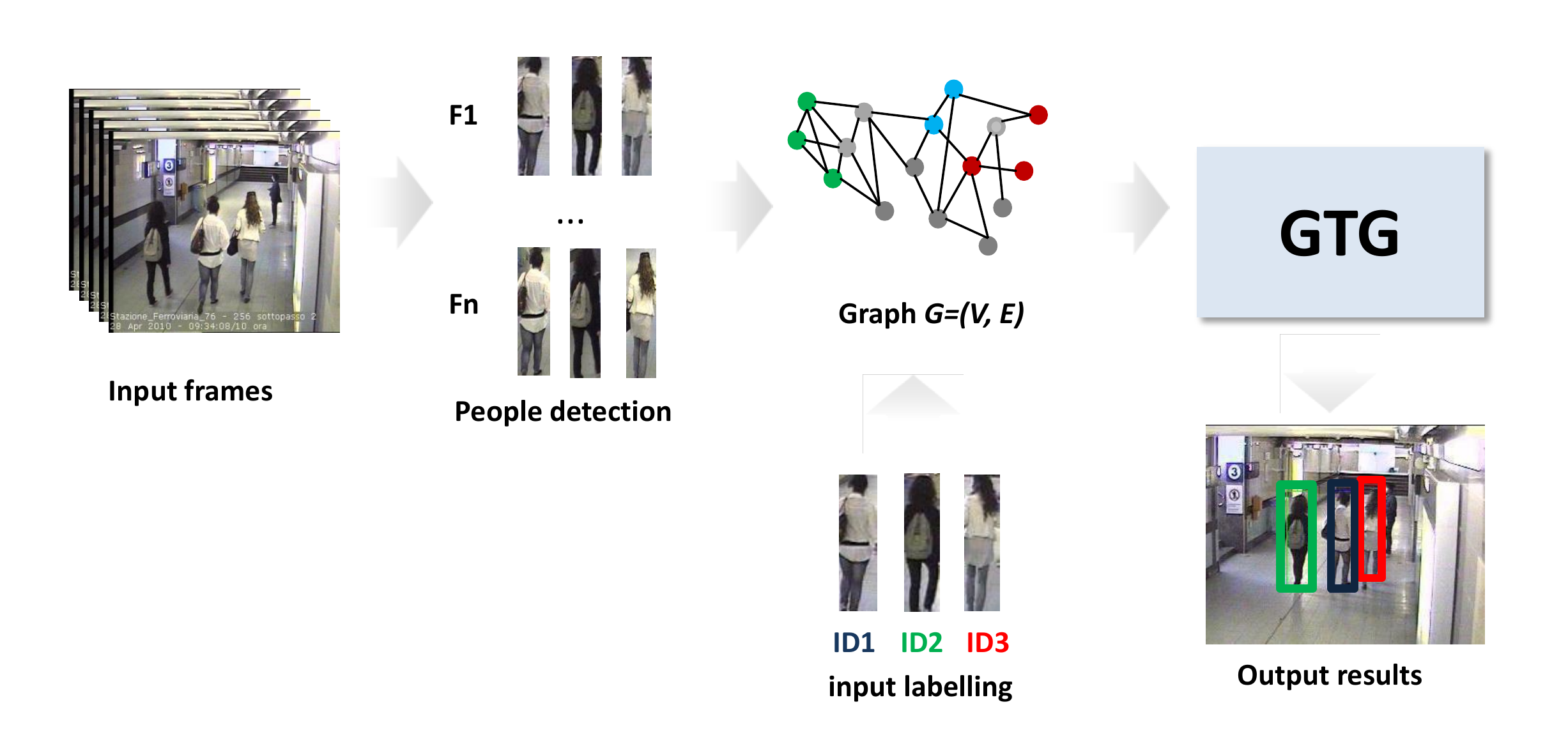} 
   \caption{Overview.}
   \label{fig:scheme}
\end{figure*}

This section describes the extension of the theoretical GTG framework presented in the previous section, to a solution for multiple people tracking and details the graph construction and the people extraction and description.
An overview of the main steps of the proposal are depicted in Fig. \ref{fig:scheme}.

We assume that the number of classes $C$ is known and that all classes are present in the labelled data \ie we assume that the initial targets estimates are given. The labels $y_i$ for the labelled points are in the set $y_i = {1, \ldots, C}$ and represent the \textit{IDs} of the different targets.

Since graph-based learning methods enforce label smoothness over the graph, it is important that the graph is built exploiting a meaningful and robust similarity measure capable of capture people similitude in patches of the same target and differentiate between different targets.
We decided to model people appearance using covariance matrix feature descriptors, \cite{tuzel05}. This representation has been adopted in multiple approaches \cite{metternich10, liu10, coppi11} because of its robustness in capturing shape, location and color information.

\subsection{Covariance representation}
\begin{figure}[tb]
\includegraphics[width=0.5\textwidth]{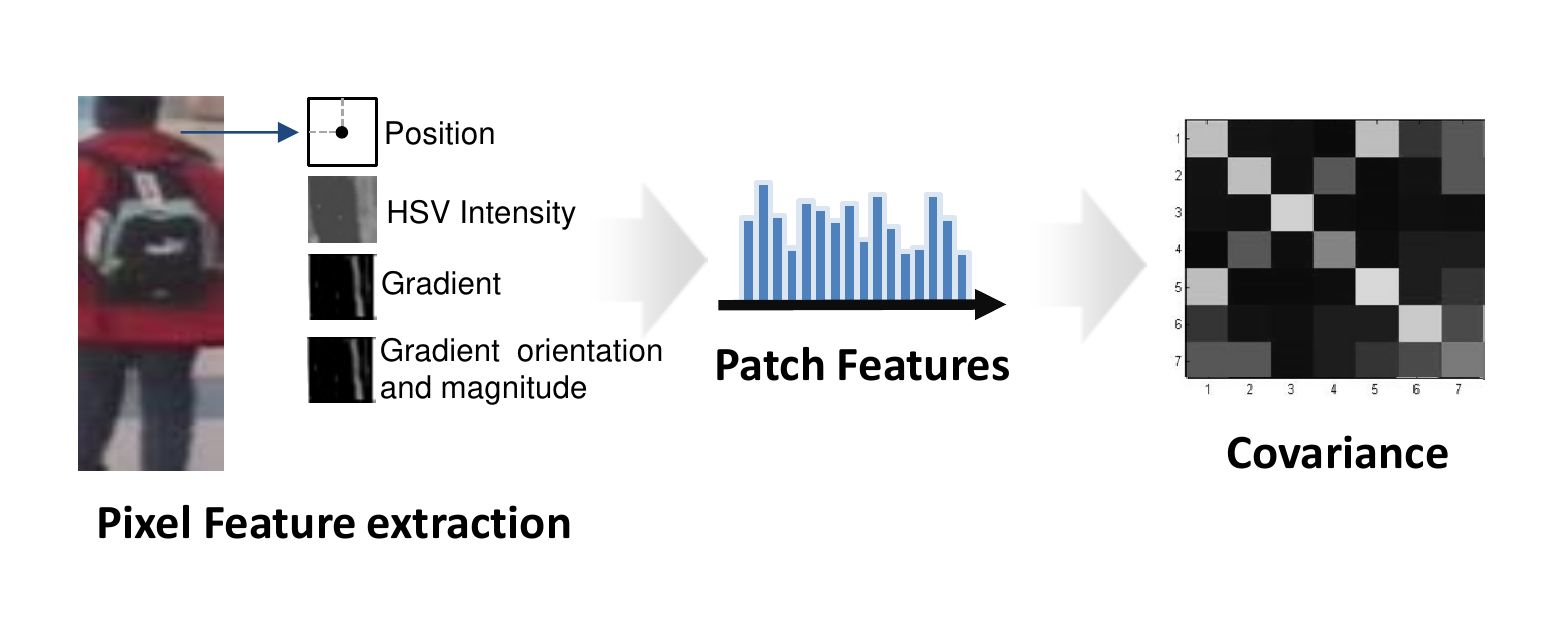}
\caption{Covariance matrix computation.}
   \label{fig:covariance}
\end{figure}

Considering $d$ different pixel features extracted from the image patches, the resulting covariance matrix $C$ is a square symmetric matrix $d \times d$ where the diagonal entries represent the variance of each feature and the non-diagonal entries represent the correlations. We decided to model each pixel within a people patch with its position $(x,y)$, its \emph{HSV} intensity values and its derivatives information. The 9-dimensional resulting feature vector $z_i$ is thus :
\begin{equation}
z_i= \left[\ x \ \ y \ \ H \ \ S \ \ V \ \ G_x \ \ G_y \ \ mag(x,y) \ \ o(x,y) \ \right]^T
\end{equation}
where $G_x$ and $G_y$ are the first order derivatives of the intensities calculated through Sobel operator w.r.t. $x$ and $y$, $mag(x,y)= \sqrt{G_x^2+G_y^2}$ and $o(x,y)=arctan\left( \frac{G_y}{G_x}\right) $ are respectively the magnitude and the angle of the first order derivatives.
The covariance $C$ of a patch of size $n = w \times h$ is then computed as $C=\dfrac{1}{n-1} \sum_{i=1}^n (z_i - \mu)(z_i - \mu)^T$ with $\mu$ the means of the vectors $z_i$. A schematic representation of the feature extraction is shown in Fig. \ref{fig:covariance}.
The choice of using the HSV color space instead of the basic RGB is because of its invariance w.r.t. scale and shift variance of light intensity.

\subsection{Graph construction}
In order to build the payoff matrix $A$, the distances between covariance matrices are necessary, and, since they do not lie on the Euclidean space, we need to use an \emph{ad-hoc} metric. Such distance metric between covariance matrices, proposed in \cite{forstner99} is equal to the sum of the squared logarithms of the generalized eigenvalues. Formally the distance between two matrices $C_i$ and $C_j$ is expressed as:
\begin{equation}
\rho (C_i, C_j) = \sqrt{\sum_{i=1}^d \ln^2 \lambda_k (C_i, C_j)}
\label{eq:cov_mat_dist}
\end{equation}
where $ {\lambda_k(C_i, C_j)}_{k=1 \ldots d} $ are the generalized eigenvalues. It is proven that $\rho$ satisfies the metric axioms, positivity, symmetry, triangle inequality, for positive definite symmetric matrices.

Assuming that our system works off-line on pre-recorded videos, the adjacency matrix $W$ is built over the complete set of people patches extracted from all the images composing the video. The pay-offs are then specified in terms of 
the normalized similarity matrix \mbox{$\widehat{W}=D^{-1/2}WD^{-1/2}$} with \mbox{$D=(d_{ii})$} being the diagonal degree matrix of $W$ whose elements are given by \mbox{$d_{ii}=\sum_j w_{ij}$} since it has been demonstrated that the normalization yield to better performance than the original similarity. 

In our particular configuration the players of the game are the different target to track and finding the Nash equilibria is equivalent to find the association of the targets in the set of images of the video stream. The


\section{Experiments}
\label{sec:experiments}
In this section we report the results obtained with the two proposed configurations on-line and off-line for people tracking and association. We evaluate our proposal on a set of video sampled from three datasets, namely:
\begin{itemize}
\item \textbf{THIS} \footnote{http://www.openvisor.org}: the videos we used in the evaluation are recorded along the platforms and underpasses of a train station and usually exhibit people walking in small groups or alone.
\item \textbf{CAVIAR} \footnote{http://homepages.inf.ed.ac.uk/rbf/CAVIAR/caviar.htm}: clips of this dataset are collected in the hallway of a shopping center in Portugal.  
\item \textbf{3DPes} \footnote{http://www.openvisor.org/3dpes.asp}: this dataset, \cite{baltieri11}, consists of videos collected at different times of the day in an university campus from different surveillance cameras. The main challenge of this dataset, originally proposed for re-identification tasks, is the large number of occlusions of the targets, we thus use it to evaluate the behaviour of our method in critical situation.
\end{itemize}

\begin{figure}[tb]
\includegraphics[width=0.45\textwidth]{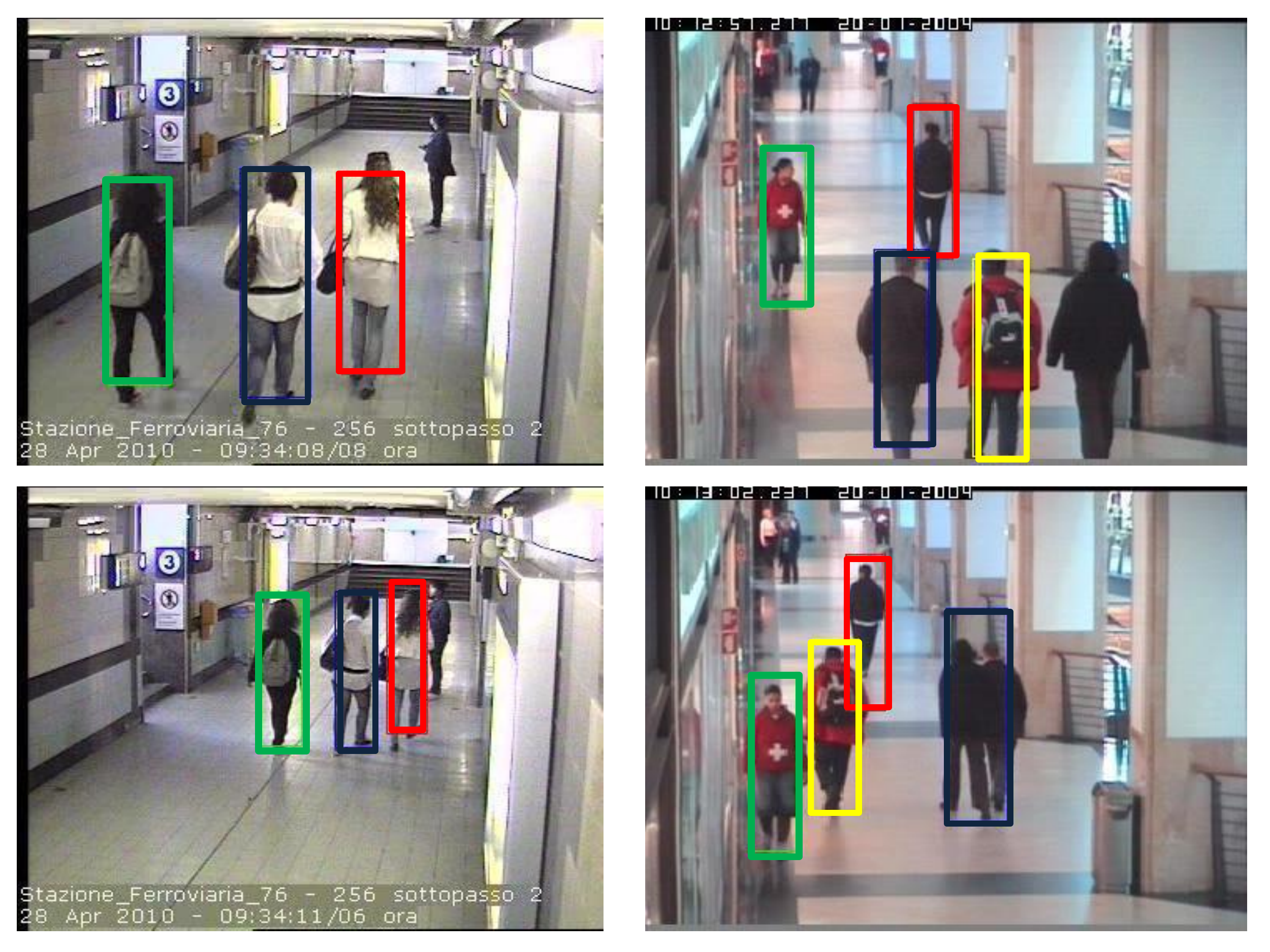} 
\caption{Examples of frames taken from THIS (left) and Caviar (right) datasets. Coloured bounding boxes show the obtained tracking results. }
   \label{fig:thiscaviar}
\end{figure}

\begin{figure}[tb]
\includegraphics[width=0.45\textwidth]{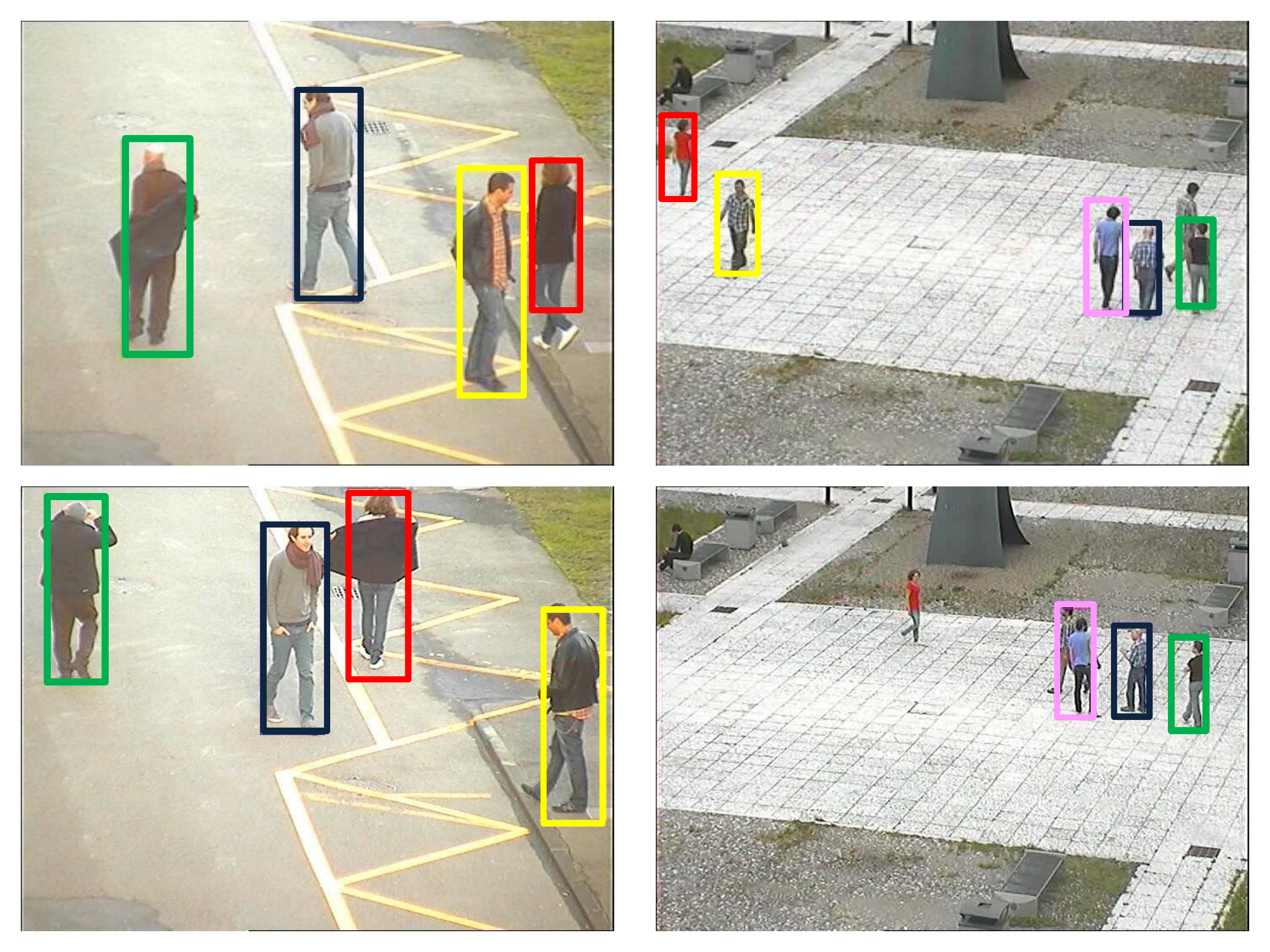} 
\caption{Examples of frames taken from the 3dPes datasets. Coloured bounding boxes show the obtained tracking results.}
   \label{fig:3dpes}
\end{figure}
Fig. \ref{fig:thiscaviar} and Fig. \ref{fig:3dpes} show some frames taken respectively from \textbf{THIS} and \textbf{Caviar} and \textbf{3dPes}.
As explained in Sec. \ref{sec:tracking} the GTG framework assumes that for each frame the patches containing the people on the scene are given, thus we initially extracted those patches using a conventional people detector based on Histogram of Oriented Gradient.
Working off-line on the people patches the problem can also be considered as one of multiple object data association, therefore we measured the performance in terms of mean object Precision and mean object Recall, where we considered a \emph{True Positive} as a patch classified correctly with its label, a \emph{False Positive} a misclassified patch (\eg the $i^{th}$ target label is assigned to a different person) and \emph{False Negative} a missing estimation (\eg the $i^{th}$ target label is non assigned in the frame even if that target is present).
In order to abstract the overall performance we also evaluate the F-measure.
 
\begin{figure*}[ht!]
   \centering \subfigcapskip = 1 pt \subfigbottomskip = 1 pt \subfigtopskip = 1 pt
   \subfigure[\label{fig:offLinePrec}]
   {\includegraphics[width=0.45\textwidth]{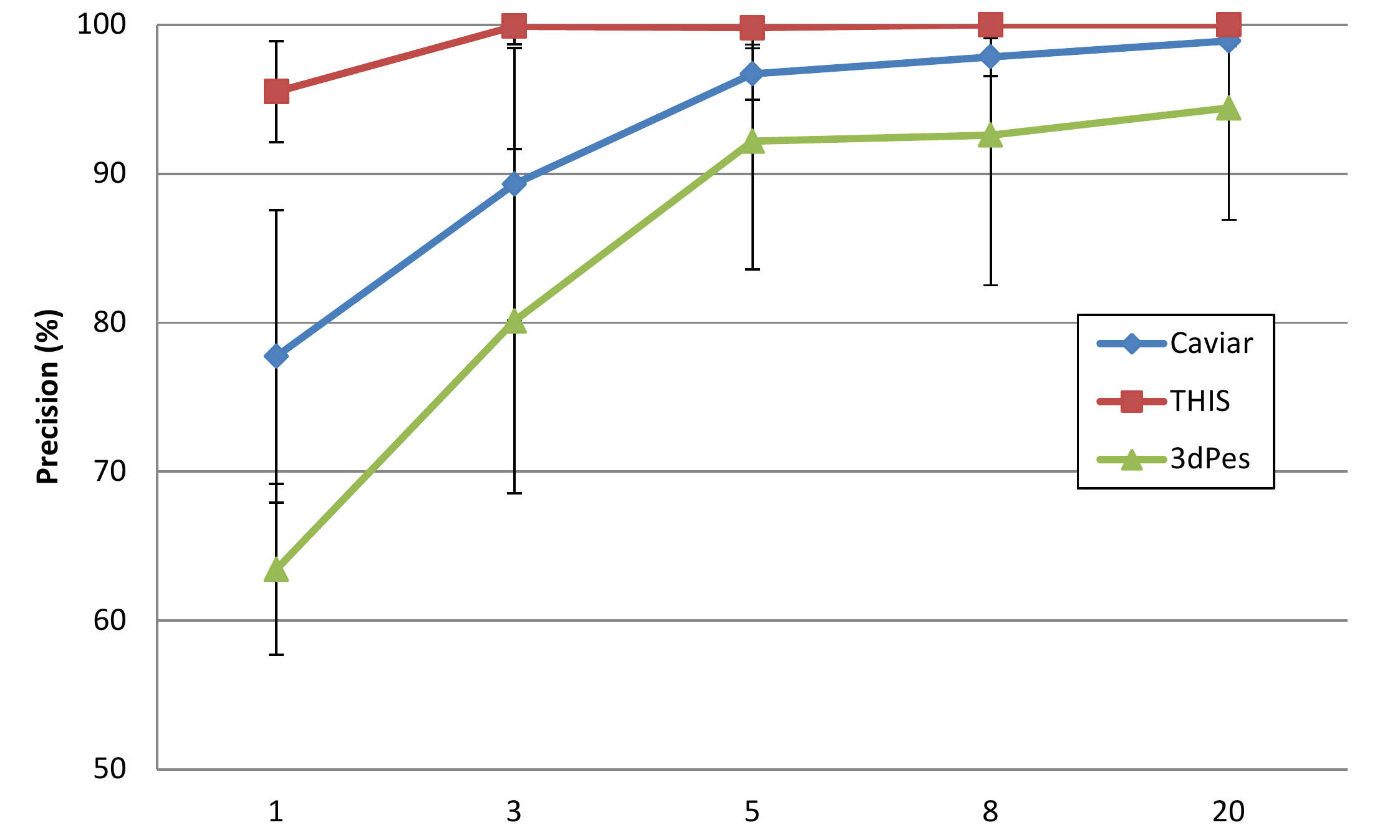}}  
   \qquad 
   \subfigure[\label{fig:offLineRec}]
   {\includegraphics[width=0.45\textwidth]{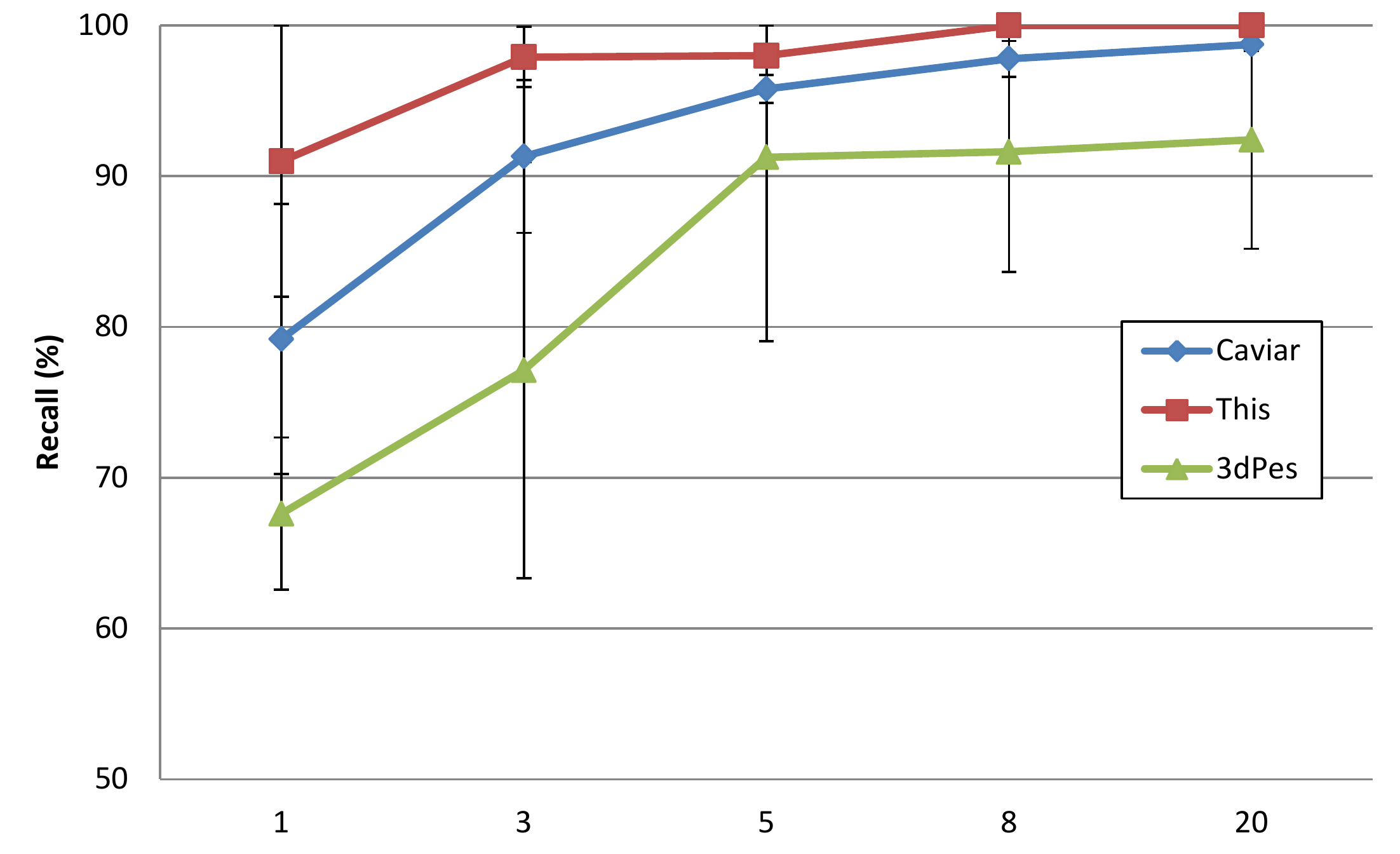}}
\caption{Results reported in terms of Precision (a) and Recall (b) varying the labelled input frames. The number of labelled frames is reported on the horizontal axis.}
   \label{fig:offLineEval}
\end{figure*}

Since the approach we proposed is off-line and relies on pre-recorded sequence, we evaluate the results varying the number of initially labelled frames. These frames were randomly chosen and the results have been averaged on 20 different executions. 
Table \ref{tab:multitarget} reports the F-measure obtained using a initial labelling of five frames on the three datasets and summarizes the averaged length of the evaluated videos in number of frames and the number of targets to track for each dataset.
Fig. \ref{fig:offLinePrec} and \ref{fig:offLineRec} show the values of mean precision and recall on the three datasets.

Clearly the obtained results improve as we increase the labelled frames as illustrated in Fig. \ref{fig:offLineEval}, but almost saturate to satisfactory values when the number of frames is fixed to 5 demonstrating good reliability without requiring a large number of labelled data. The results also show how increasing the number of labelled frames also increase the stability of the solution while labelling only one or three frames, despite the precision and recall reach adequate levels, the standard deviation is large. 
The best performing dataset is \textbf{THIS} reaching the 100 \% of precision and accuracy,  but results are good even on challenging dataset such as \textbf{3dPes}, where the number of target is higher and the targets are not always correctly identified by the detector due to occlusions, though on the \textbf{3dPes} dataset the standard deviation of the results is higher. 
We would like to highlight that the video length reported in Table \ref{tab:multitarget} is an average on different sequences, but, especially with the \textbf{3dPes} dataset we stressed the algorithm increasing the number of frames from  200 to 430 and even with the highest number of frames the results were satisfactory with the F-measure of roughly the 90\%.
Few obtained tracking results are proposed in Fig. \ref{fig:thiscaviar} and Fig. \ref{fig:3dpes}.

\begin{table}[htb]
  \centering
  \caption{Multitarget tracking performances.}
    \begin{tabular}{c|rrrr}
    \toprule
    \textbf{Dataset} &  \textbf{\# Frames} & \textbf{\# Targets} & \textbf{F-measure} \\
    \midrule
    \textbf{THIS} & 109  & 3 & 0.99 \\
    \midrule
    \textbf{CAVIAR} & 140  & 4 & 0.96 \\
    \midrule
    \textbf{3dPes} & 280  & 5 & 0.92 \\
    \bottomrule
    \end{tabular}
  \label{tab:multitarget}
\end{table}
To further the evaluation we also compare our approach with the other tracking methods based on graph transduction. At this aim we recast our solution as one of single people tracking, labelling the input samples $x_i$ with $y_i = +1$ when they correspond to the target and $y_i = -1$ otherwise. Results are reported in Tab. \ref{tab:singletarget}, in particular we evaluate the GTG framework against the seminal Transductive Learning Tracker (TLT) of \cite{zha10} and the work presented in \cite{coppi11}, that proposed a single target Transduction Tracking (TT NoUpd) and also an improved version where the labelled data point are iteratively updated with an evolutionary spectral clustering (TT SpUpd). 
Even if we should specify that all the other methods work iteratively on-line on each frame of the video, our solution has been proven to be reliable even in a single target configuration and outperforms the other methods. 
The reason of the improvement over the state-of-the-art is justified both by the robustness of the GTG framework as already demonstrated in \cite{erdem12} for other classification tasks, and by the fact that working with graphs with more nodes increase the possible paths among nodes representing the target patches in different frames. In other words this overcomes the inevitable errors of the on-line methods when the people detection are imprecise in the evaluated frames.
\begin{table}[htb]
  \centering
  \caption{Comparison with the single target tracking method proposed in \cite{coppi11}.}
    \begin{tabular}{c|rrrr}
    \toprule
    \textbf{Dataset} & \textbf{Method} & \textbf{Precision} & \textbf{Recall} & \textbf{F-measure} \\
    \midrule
    \multirow{4}{*}{\textbf{THIS}} 
    	  & {\em \textbf{GTG}} & 1.00  & 0.85 & 0.92 \\
    	  & {\em \textbf{TLT}} & 0.76  & 0.92 & 0.83 \\
    	  & {\em \textbf{TT NoUpd}} & 0.80  & 0.93 & 0.86\\   	  
          & {\em \textbf{TT SpUpd}} & 0.97  & 0.95  & 0.96 \\    
          \midrule
    \multirow{4}{*}{\textbf{CAVIAR}} 
    	  & {\em \textbf{GTG}} & 0.90  & 0.92 & 0.91 \\
    	  & {\em \textbf{TT NoUpd}} & 0.68  & 0.91 & 0.78 \\   	  
          & {\em \textbf{TT SpUpd}} & 0.78  & 0.90  & 0.84\\
          & {\em \textbf{TLT}} & 0.96  & 0.94 & 0.95 \\
          \midrule
    \multirow{4}{*}{\textbf{3dPes}} 
    	  & {\em \textbf{GTG}} & 0.87  & 0.99 & 0.93 \\
    	  & {\em \textbf{TT NoUpd}} & 0.45  & 0.44 & 0.44\\   	  
          & {\em \textbf{TT SpUpd}} & 0.63  & 0.66 & 0.64\\
          & {\em \textbf{TLT}} & 0.86  & 0.83  & 0.84 \\
    \bottomrule
    \end{tabular}
  \label{tab:singletarget}
\end{table}

\section{Conclusion}
\label{sec:conclusion}
In this paper we proposed a method for multitarget people tracking based on a Graph Transduction Game framework.
The GTG considers the different targets to track as players of a game, and estimates the missing labels by finding the Nash equilibria over the set of possible strategies, where the pay-offs of the different strategies are proportional to the similarity among people patches. We extracted people patches by using a HOG based detector and we represented their appearance with a covariance matrix.

The method has been evaluated over three different video surveillance datasets and obtained good results even with very few labelled initial frames. Our proposal also outperforms similar methods when considered in a single target configuration.

Given the promising results we obtained future work includes the extension to longer video sequence in order to test the robustness of the system also for long term tracking. 
We would like also to evaluate a on-line approach and insert mechanism to handle the insertion of new targets, or in other words to handle a variable number of classes $C$. Always in the framework of an on-line label estimation we would like to test the possibility of avoid the initial step of people detection, and instead, to use a sliding window approach to extract the patches, such as the method is independent from the performance of the detector.

\bibliographystyle{IEEEtran}
\bibliography{GTMTT}
\vfill

\end{document}